\documentclass[10pt,twocolumn,letterpaper]{article}
\usepackage[pagenumbers]{cvpr} 

\usepackage{graphicx}
\usepackage{amsmath}
\usepackage{amssymb}
\usepackage{booktabs}
\usepackage{url}
\usepackage{multirow}
\usepackage{multicol}
\usepackage{algpseudocode}
\usepackage{listings}
\usepackage{tablefootnote}
\usepackage[dvipsnames]{xcolor}
\usepackage{enumitem}
\usepackage{algorithm}
\usepackage{algpseudocode}
\usepackage{sidecap,wasysym,array,tabularx,caption}

\algrenewcommand{\Return}{\State\algorithmicreturn~}
\usepackage{colortbl}
\usepackage[normalem]{ulem}
\usepackage{relsize,amsfonts}
\usepackage{import}
\usepackage[accsupp]{axessibility}
\definecolor{mygray}{gray}{.9}

\newcommand{\encoderparam}{$\phi_{\mathcal{E}}$}
\newcommand{\decoderparam}{$\phi_{\mathcal{D}}$}
\newcommand{\discriminatorparam}{$\phi_{\Gamma}$}
\newcommand{\encoder}{$\mathcal{F}_{\phi_{\mathcal{E}}}$}
\newcommand{\decoder}{$\mathcal{F}_{\phi_{\mathcal{D}}}$}
\newcommand{\discriminator}{$\mathcal{F}_{\phi_{\Gamma}}$}
\newcommand{\fasking}{$\mathcal{F}_{\phi_{f}}$}

\usepackage[pagebackref,breaklinks,colorlinks]{hyperref}

\usepackage[capitalize]{cleveref}
\crefname{section}{Sec.}{Secs.}
\Crefname{section}{Section}{Sections}
\Crefname{table}{Table}{Tables}
\crefname{table}{Tab.}{Tabs.}

\begin{document}

\title{MARLIN: \textbf{M}asked \textbf{A}utoencoder for facial video \textbf{R}epresentation \textbf{L}earn\textbf{IN}g}

\def\MH#1{{\bf [} {\it \color{red} {#1}}{\bf ]}.}

\author{Zhixi Cai$^1$, Shreya Ghosh$^{1,2}$, Kalin Stefanov$^1$, Abhinav Dhall$^{1,3}$, Jianfei Cai$^1$,\\ Hamid Rezatofighi$^1$, Reza Haffari$^1$, Munawar Hayat$^1$\\
$^1$Monash University, $^2$ Curtin University, $^3$ Indian Institute of Technology Ropar\\
\small \texttt{\{zhixi.cai,kalin.stefanov,jianfei.cai,hamid.rezatofighi,gholamreza.haffari,}\\ \small \texttt{munawar.hayat\}@monash.edu,shreya.ghosh@curtin.edu.au,abhinav@iitrpr.ac.in}}

\maketitle

\begin{abstract}
This paper proposes a self-supervised approach to learn universal facial representations from videos, that can transfer across a variety of facial analysis tasks such as Facial Attribute Recognition (FAR), Facial Expression Recognition (FER), DeepFake Detection (DFD), and Lip Synchronization (LS). Our proposed framework, named \textbf{MARLIN}, is a facial video masked autoencoder, that learns highly robust and generic facial embeddings from abundantly available non-annotated web crawled facial videos.
As a challenging auxiliary task, MARLIN reconstructs the spatio-temporal details of the face from the densely masked facial regions which mainly include eyes, nose, mouth, lips, and skin to capture local and global aspects that in turn help in encoding generic and transferable features. 
Through a variety of experiments on diverse downstream tasks, we demonstrate MARLIN to be an excellent facial video encoder as well as feature extractor, that performs consistently well across a variety of downstream tasks including FAR (1.13\% gain over supervised benchmark), FER (2.64\% gain over unsupervised benchmark), DFD (1.86\% gain over unsupervised benchmark), LS (29.36\% gain for Frechet Inception Distance), and even in low data regime.
Our code and models are available at \href{https://github.com/ControlNet/MARLIN}{https://github.com/ControlNet/MARLIN}.

\end{abstract}

\section{Introduction}
\label{sec:intro}

Facial analysis tasks~\cite{guoMSCeleb1M2016,zhuCelebVHQ2022,tolosanaDeepfakes2020,kadamSurvey2021} provide essential cues for human non-verbal behavior analysis, and help unfold meaningful insights regarding social interaction~\cite{haughFace2009}, communication~\cite{jackHuman2015}, cognition~\cite{storrsUnsupervised2021} with potential applications in Human-Computer Interaction (HCI) and Affective Computing domains. Recently, we have witnessed significant progress in deep neural network models to solve facial analysis tasks such as Facial Attribute Recognition (FAR)~\cite{guoMSCeleb1M2016,zhuCelebVHQ2022}, Facial Expression Recognition (FER)~\cite{liDeep2022a}, DeepFake Detection (DFD)~\cite{tolosanaDeepfakes2020}, and Lip Synchronization (LS)~\cite{kadamSurvey2021}. While these deep models can achieve remarkable performance, they often require large-scale annotated datasets, which is not only a resource-expensive and time-consuming process but also infeasible for some applications requiring domain expertise for annotation (e.g. FER). 

\begin{figure}
    \centering
    \includegraphics[width = \linewidth]{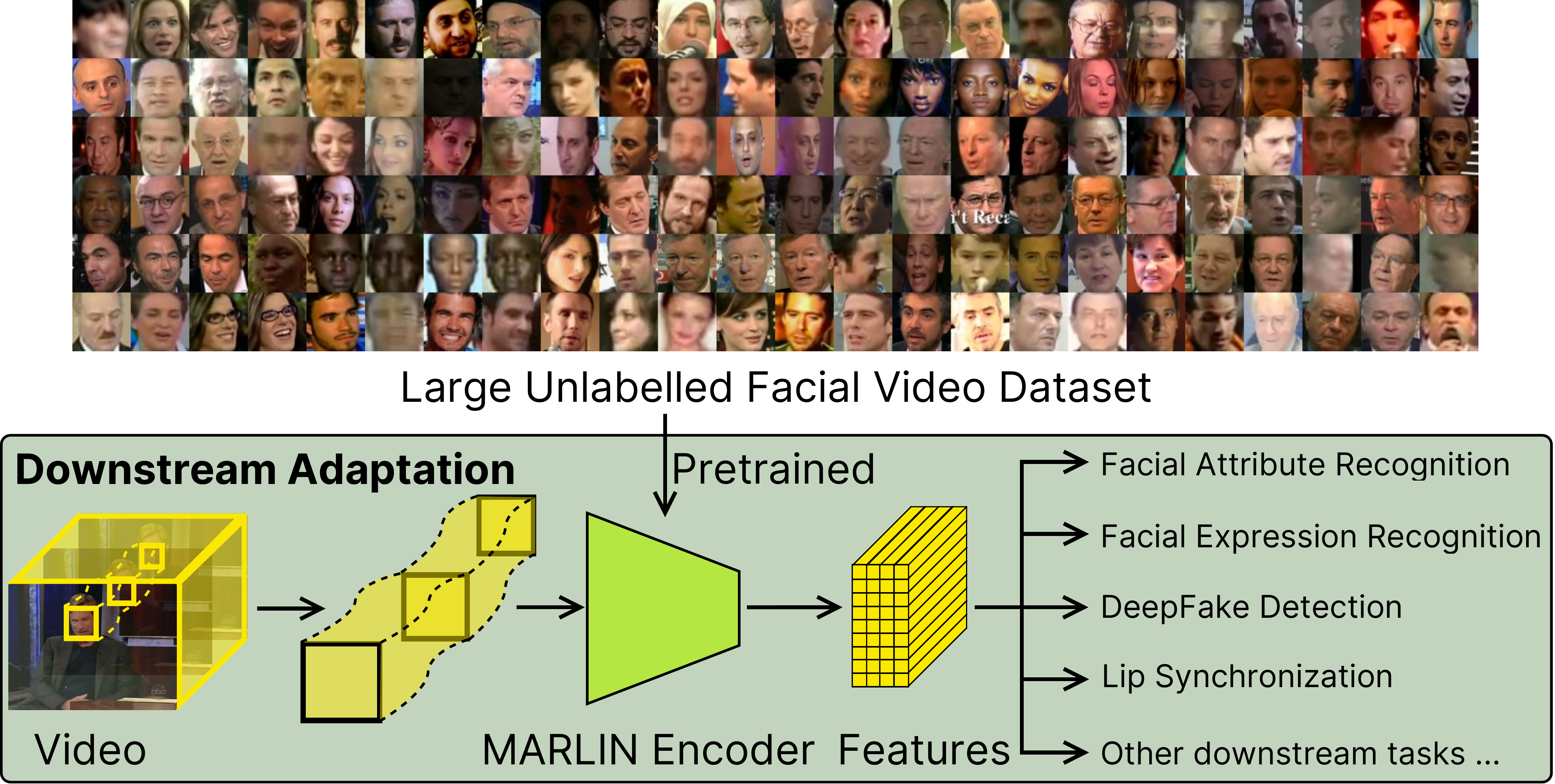}
    \caption{Overview of the proposed Masked Autoencoder for facial Representation LearnINg aka MARLIN. MARLIN aims to learn a universal facial representation from abundantly available non-annotated facial video data.}
    \label{fig:teaser}
    \vspace{-1mm}
\end{figure}

To this end, self-supervised pre-training ~\cite{devlinBERT2019, heMasked2022, tongVideoMAE2022} has lately emerged as an effective strategy to address the limitations of fully supervised methods, as it enables generic representation learning from non-annotated data, that can then be transferred across tasks having limited labels. For images of natural scenes and objects, self-supervised learning approaches using self-distillation~\cite{caronEmerging2021}, contrastive-learning~\cite{chenBig2020,chenSimple2020}, solving pre-text tasks such as jigsaw puzzle~\cite{norooziUnsupervised2016}, and more recently autoencoding~\cite{heMasked2022,tongVideoMAE2022} have even outperformed the supervised learning approaches. 

Despite the promises offered by these self-supervised methods in learning scalable and generic representations for natural scene images and videos, these have not yet been investigated for learning representations from facial video data. Facial representation learning requires tracking of  fine-grained face specific details which might not be perfectly captured by linear tube masking~\cite{tongVideoMAE2022}. Until now, most of the existing approaches associated with facial analysis tasks are highly specialized and develop task-specific models trained in a fully supervised manner \cite{schroffFaceNet2015,parkhiDeep2015,korshunovDeepFakes2018}, with very few recent efforts towards learning generic \textit{image-based} facial encoding \cite{zhengGeneral2022,bulatPretraining2022}. These closely related works~\cite{bulatPretraining2022,zhengGeneral2022} either focus on exploring training dataset properties in terms of size and quality~\cite{bulatPretraining2022} or performing pre-training in visual-linguistic way~\cite{zhengGeneral2022}. These works~\cite{bulatPretraining2022,zhengGeneral2022} are hard to scale since they use static image-level facial information and the image-caption pairs are highly associated with context information rather than face.

In this paper, our goal is to learn \textit{universal} and \textit{task-agnostic} representations in a self-supervised manner for face-related downstream tasks (see Fig.~\ref{fig:teaser}). For this purpose, we employ a masked autoencoder~\cite{heMasked2022,tongVideoMAE2022} with a facial-guided masking strategy that learns to reconstruct spatio-temporal details of a face from densely masked facial regions using non-annotated videos. Unlike existing approaches for natural scene videos ~\cite{tongVideoMAE2022}, where the tube-masking is initialized with a static part of the video without any semantic information, our approach dynamically tracks face and then develops a facial part-guided tube masking strategy using an off-the-shelf face parser i.e. FaceXZoo~\cite{wangFaceXZoo2021}. Thus, we pose a more challenging task that encourages the model to learn spatio-temporal representations to cover local as well as global information.
Inspired by prior works~\cite{radfordUnsupervised2016, donahueLarge2019} showing high-quality reconstruction results along with rich and generic latent features, we incorporate adversarial loss on top of masked encoding to enhance reconstruction quality. Our experimental results show that our proposed framework, MARLIN, learns highly generic facial encoding that scale and transfers well across diverse facial analysis tasks such as FER, DFD, FAR, and LS and achieve favorable performance gain \wrt state-of-the-art benchmarks. In summary, our main contributions are:

\begin{itemize}[topsep=1pt,itemsep=0pt,partopsep=1ex,parsep=1ex,leftmargin=*]
    \item We propose, MARLIN, a \textit{universal} and \textit{task-agnostic} facial encoder that learns robust and transferable facial representation from abundantly available non-annotated web-crawled facial videos in a self-supervised fashion.
    
    \item As a challenging auxiliary task, we propose to reconstruct the spatio-temporal details of the face from the densely masked facial regions. The proposed facial region-guided tube masking (aka \textit{Fasking}) strategy aims to learn local and global aspects from facial videos which in turn help encode generic and transferable features. 
    
    \item Through extensive quantitative and qualitative analysis, we show that MARLIN learns rich, generic, transferable, and robust facial representation, that performs consistently well across a variety of downstream tasks including FAR (1.13\% gain over supervised benchmark), FER (2.64\% gain over unsupervised benchmark), DFD (1.86\% gain over unsupervised benchmark), LS (29.36\% gain for Frechet Inception Distance)
    and even in few shot settings.
    
\end{itemize}

\section{Related Work}

\noindent \textbf{Masked AutoEncoder.} Masked autoencoder learns robust and transferable representations based on the hypothesis of reconstruction of the masked region. Masked autoencoding is motivated by context encoders~\cite{pathakContext2016} and denoising encoders~\cite{vincentStacked2010}. After success of BERT~\cite{devlinBERT2019} based masking, the vision community has also explored different design choices of masked auto encoding such as pixel level masking ~\cite{chenGenerative2020,heMasked2022,xieSimMIM2022}, token level masking~\cite{el-noubyAre2021} and deep feature based masking~\cite{weiMasked2022,baevskidata2vec2022}, using vision Transformers \cite{khanTransformers2021, naseerIntriguing2021}. Similarly, for modeling spatio-temporal patterns of the input data, masked motion modelling~\cite{sunVideo2022} and tube masking~\cite{tongVideoMAE2022} strategies have been incorporated recently. Along this line, MARLIN masks and reconstructs domain-specific facial parts to learn universal facial representation.

\begin{table}[t]
    \centering
    \caption{\textbf{Facial Analysis Tasks.} Overview of different face related tasks and relevant datasets down the lane.}
    \scalebox{0.75}{
    \begin{tabular}{l||c c c c c}
    \toprule[0.4mm]
    \rowcolor{mygray}  Datasets & \# Samples & Env. &  Fmt. & Task & Year\\ \hline \hline
    LFW~\cite{huangLabeled2008} & 13,233 & Wild & Img. & Identification & 2008\\
    VGG-FACE~\cite{parkhiDeep2015} & 2.6M & Wild &  Img. & Identification & 2015\\
    CelebA~\cite{liuDeep2015} & 202,599 & Wild &  Img. & Attributes & 2015\\\hline
     
   YouTubeFace~\cite{wolfFace2011}  & 3,425 & Wild & Vid & Identification & 2011\\
   LRS2~\cite{chungLip2017} & 144,482 & Wild &  Vid & Lip Sync. & 2017\\
   CelebV~\cite{wuReenactGAN2018}  & 5 & Wild &  Vid & Reenact & 2018\\
   CMU-MOSEI~\cite{zadehMultiattention2018a} & 23,453  & Wild &  Vid & Emo, Senti & 2018\\
   FaceForensics++~\cite{rosslerFaceForensics2019} & 1,004 & Wild &  Vid & DeepFake & 2019\\
   VoxCeleb2~\cite{chungVoxCeleb22018} & 150,480 & Wild & Vid & Speaker & 2018\\
   CelebV-HQ~\cite{zhuCelebVHQ2022} & 55,666 & Wild &  Vid & Attribute & 2022\\ \bottomrule[0.4mm]
    \end{tabular}}
    \vspace{-2mm}
    \label{tab:datasets}
\end{table}

\noindent \textbf{Facial Representation Learning.} 
Till date, most of the existing facial analysis approaches are conducted in a task-specific way with fully supervised manner \cite{schroffFaceNet2015,parkhiDeep2015,korshunovDeepFakes2018} on manually annotated data to enhance performance. Any state-of-the-art model's performance on benchmarked datasets is impacted by the quality and quantity of annotated data used during training.  Tab.~\ref{tab:datasets} shows an overview of the task-specific large-scale facial image or video datasets that have been curated over the past decade~\cite{adjabiPresent2020} to facilitate research in Face Verification (LFW~\cite{huangLabeled2008}, MS-celeb-1M~\cite{guoMSCeleb1M2016}, VGG-FACE~\cite{parkhiDeep2015}, VGGFace2~\cite{caoVGGFace22018}), Facial Attribute Recognition(CelebA~\cite{liuDeep2015}, CelebV-HQ~\cite{zhuCelebVHQ2022}), Facial Emotion Recognition (CMU-MOSEI~\cite{zadehMultiattention2018a}), DeepFake Detection (FF++~\cite{rosslerFaceForensics2019}) and Lip Synchronization (LRS2~\cite{chungLip2017}). However, data curation encounters several challenges such as requirements of specialized hardware (e.g. for FER and action unit data), the discrepancy in data distribution that prevent merging of multiple datasets~\cite{bulatPretraining2022}, and most importantly time consuming and resource expensive annotation process. To eliminate these drawbacks, some of the existing approaches~\cite{choiStarGAN2020,yanVideoGPT2021,yuGenerating2022} adopt data augmentation strategy via image or video synthesis as the surge in face generation technology fueled by Generative Adversarial Network (GAN)~\cite{choiStarGAN2020,yanVideoGPT2021,yuGenerating2022,skorokhodovStyleGANV2022} and other generation techniques~\cite{haoGANcraft2021,chanPiGAN2021} aids realistic face generation even with the control over facial attributes. These generation techniques add variation in training set quantitatively, but in some cases it still lags in qualitative aspects due to domain specific inconsistency and more importantly high network complexity.

To this end, there are very few recent works that aim to learn \textit{image-based} task specific facial encoding with limited supervision~\cite{zhengGeneral2022,bulatPretraining2022,anejaGeneralized2020,zhuangFewShot2021,browatzki3FabRec2020,shuLearning2021,zhuangFewShot2021,zhengGeneral2022}. The most closely related existing works~\cite{bulatPretraining2022,zhengGeneral2022} either focus on exploring training dataset properties in terms of size and quality~\cite{bulatPretraining2022} or performing pre-training in visual-linguistic way~\cite{zhengGeneral2022}. These works~\cite{bulatPretraining2022,zhengGeneral2022} are hard to scale since they use static image level facial information and the image-caption pairs are highly correlated with context information rather than face. In this work, we aim to develop a generic, universal, and task-agnostic facial encoder that learns from web-crawled non-annotated data. Our experimental analysis shows that MARLIN can align the latent space manifold to any desired downstream task specific label space. Thus, MARLIN has the capability to act as a strong facial encoder or feature extractor in many low-resource real world applications.

\begin{figure*}
    \centering
    \includegraphics[width =\linewidth]{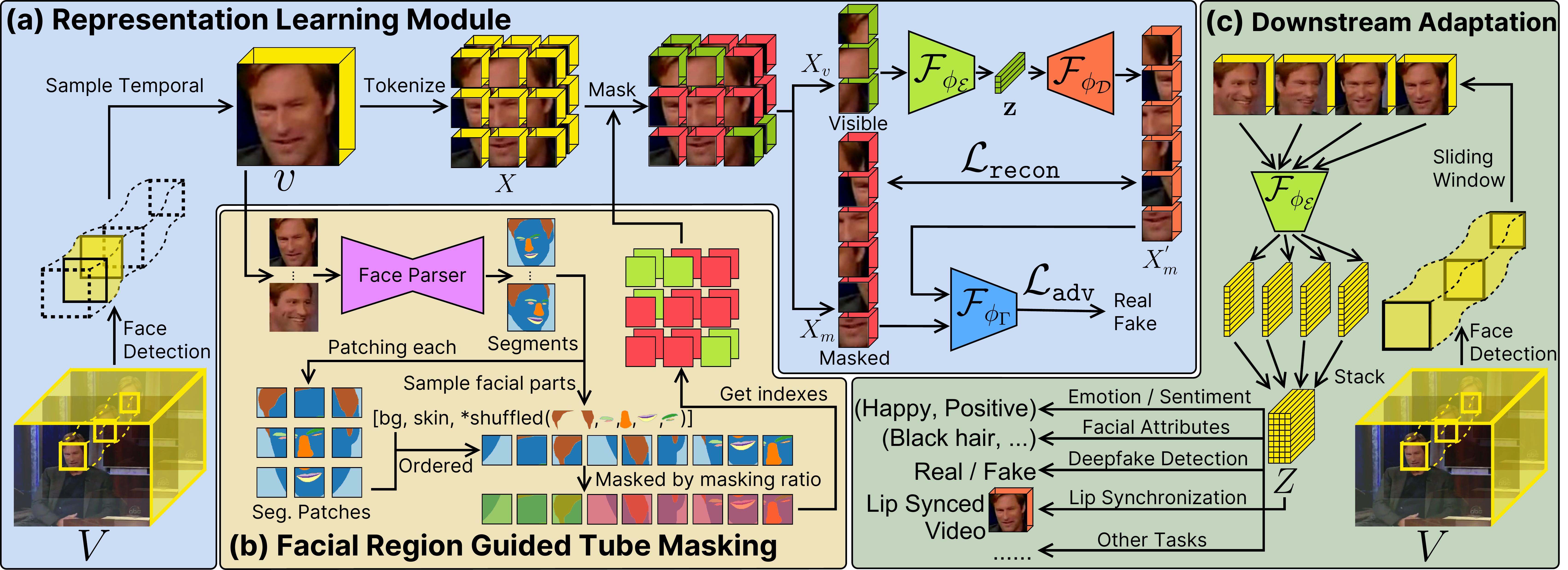}
    \caption{\textbf{Architectural overview of MARLIN (Best viewed in color).} MARLIN mainly consists of (a) Representation Learning Module, (b) Facial Region guided Tube Masking, and (c) Downstream Adaptation. (a) \textit{Representation Learning Module:} MARLIN learns the facial representation from the unlabeled, web crawled video data in a self-supervised fashion (highlighted in \textcolor{Cerulean}{Blue}). (b) \textit{Facial Region guided Tube Masking:} With the aid of facial region guided tube masking (highlighted in \textcolor{Apricot}{Yellow}), MARLIN gets joint spatio-temporal attention which in turn facilitates downstream performance. The Face guided tube masking strategy injects domain knowledge into the pipeline. (c) \textit{Downstream Adaptation:} For facial task specific downstream adaptation, MARLIN utilizes Linear Probing (LP) and Fine-Tuning (FT) to show the robustness, generalizability, and transferability of the learned feature (highlighted in \textcolor{OliveGreen}{Green}).}
    \label{fig:pipeline}
    \vspace{-3mm}
\end{figure*}

\section{MARLIN}\label{sec:marlin}
Our objective is to learn robust and transferable universal facial representation from abundantly available non-annotated facial video data~\cite{wolfFace2011}. 
If we think holistically, face specific tasks involve two different aspects: a) facial appearance related attributes such as parts of the face (nose, eyes, lips, hair, etc.), facial shape and texture which mainly need spatial investigation; and b) facial action such as emotion, Facial Action Coding System (FACS), lip synchronization which requires temporal information. Thus, spatio-temporal modeling is highly desirable in order to learn strong, robust, and transferable representation. To this end, our proposed framework, MARLIN, adopts a facial region guided masking strategy which poses a challenging auxiliary reconstruction task for self supervised representation learning (See Fig.~\ref{fig:pipeline}). 
To facilitate learning from masked auto-encoder, we mainly choose the YouTube Faces~\cite{wolfFace2011} dataset that uses web-crawled facial videos from YouTube having variation in terms of different real life conditions. 

\begin{algorithm}[tb]
\caption{Facial-region Guided Masking Procedure}
\label{alg:face}
\begin{algorithmic}[1]
\Require{$v\in \mathbb{R}^{(C\times T \times H \times W)}$, $\mathbf{r}$}
\State  \texttt{seg\_map} $\gets \texttt{FaceXZoo}(v)$ \Comment{Face-Parsing,  \texttt{seg\_map}$\in$\{\texttt{background,skin,left-eye,
right-eye,nose,mouth,hair}\}}
\State $\mathbb{P} = $\{\texttt{left-eye, right-eye, nose, mouth, hair}\} \Comment{Prioritize Regions}
\State $\mathbf{k} = \frac{T}{\mathbf{t}} \times \frac{H}{\mathbf{h}} \times \frac{W}{\mathbf{w}}$ \Comment{\# of tokens for each $v$ (3D cube tokens have dimension of $\mathbf{t}\times \mathbf{h}\times \mathbf{w}$ each)}
\State $\mathbf{n} \gets \mathbf{r} \times \mathbf{k}$ \Comment{Number of masked tokens}
\State $\tilde{X}_v \gets \{\}$ \Comment{Initialize visible tokens}
\State\texttt{patches} =\{\texttt{background,skin,*\texttt{shuffle}$(\mathbb{P})$}\} \Comment{Ordered list}
\For{\texttt{patch} in \texttt{patches}}
\State $\tilde{X}_v \gets \{\texttt{patch}\}$
\If{$\mathbf{len}(\tilde{X}_v) == (\mathbf{k}-\mathbf{r})$} 
\State \texttt{break}
\EndIf
\EndFor
\State $\tilde{X}_m \gets \tilde{X} - \tilde{X}_v$ \Comment{$\tilde{X}$ is all tokens from $v$}
\end{algorithmic}
\end{algorithm}

\subsection{Facial Representation Learning}
\noindent \textbf{Preliminaries.} MARLIN consists of an encoder (\encoder), decoder (\decoder) and discriminator (\discriminator) with embedding parameters \encoderparam, \decoderparam\ and \discriminatorparam, respectively. Given a training dataset $\mathbb{D} = \{V_i\}_{i=1}^{N}$ where $N$ is the number of videos in the dataset and $V \in \mathbb{R}^{C\times T_0\times H_0 \times W_0}$ ($C$, $T_0$, $H_0$, $W_0$ are channel, temporal depth, height and width of the raw video, respectively). From the raw input video $V$, we track and crop the facial regions~\cite{wangFaceXZoo2021} followed by random temporal sampling represented as $v \in \mathbb{R}^{(C\times T \times H \times W)}$ ($T$, $H$, $W$ are the modified temporal depth, height, and width of the derived video clip, respectively). The derived video clip $v$ is further mapped to ($\mathbf{k}-\mathbf{n}$) visible and $\mathbf{n}$ masked tokens denoted as $\{\tilde{X}_v \in \mathbb{R}^{(\mathbf{k}-\mathbf{n})\times \mathbf{e}}, \tilde{X}_m\in \mathbb{R}^{\mathbf{n}\times \mathbf{e}}\}$ by facial-region guided masking strategy (\fasking) with a pre-defined masking ratio $\mathbf{r}=\frac{\mathbf{n}}{\mathbf{k}}$. Here, $\mathbf{e}$ is the embedding dimension and $\mathbf{k}$ is the total number of tokens derived from $v$, i.e. $\mathbf{k} = \frac{T}{\mathbf{t}} \times \frac{H}{\mathbf{h}} \times \frac{W}{\mathbf{w}}$, given 3D cube tokens have dimension of $\mathbf{t}\times \mathbf{h}\times \mathbf{w}$ each. Thus, MARLIN injects facial region specific domain knowledge in the aforementioned token space to guide the representation learning via masking. 

The visible tokens $\tilde{X}_v$ are mapped to the latent space $\mathbf{z}$ by the following mapping function $\mathcal{F}_{\phi_{\mathcal{E}}}:\tilde{X}_v \to \mathbf{z}$. The latent space feature $\mathbf{z}$ is further fed to the decoder $\mathcal{F}_{\phi_{\mathcal{D}}}$ which reconstruct $\mathbf{z}$ to the $\mathbf{n}$ masked tokens $X_m^{'}$ by the following mapping $\mathcal{F}_{\phi_{d}}:\mathbf{z} \to X^{'}_m$. In the decoder, the corresponding visible and masked 3D cubes contain the flatten raw pixels denoted as $\mathbf{e} = \mathbf{Cthw}$.
In brief given the visible tokens $\tilde{X}_v$, we reconstruct the masked tokens by the following function:  
\vspace{-0mm}
\begin{equation}
X_m^{'} = \mathcal{F}_{\phi_{\mathcal{D}}} \circ \mathcal{F}_{\phi_{\mathcal{E}}}(\tilde{X}_v)
\vspace{-0mm}
\end{equation} 

Reconstructing spatio-temporal facial patterns from raw pixels is quite challenging, we deploy a discriminator \discriminator with the adversarial training for better synthesis.

\subsection{Self-Supervised Representation Learning.}
The self supervised pre-training strategy of MARLIN consists of three main components described below:

\noindent \textbf{a) Facial-region Guided Tube Masking (Fasking).} 
In order to capture spatio-temporal correspondence, we have deployed facial region specific tube masking strategy following~\cite{tongVideoMAE2022}. We dynamically track and mask facial components across temporal axis for each spatio-temporal cube. Our facial regions based tube-masking strategy ensures the same facial region is masked throughout the temporal cube, thus posing a challenging reconstruction task and promoting learning local and global facial details (See Alg.~\ref{alg:face}). As the masked spatio-temporal cubes look like deformable bending tubes, we termed it as \textit{Facial region-guided tube masking} aka \textit{Fasking}.

We begin with face parsing using FaceXZoo~\cite{wangFaceXZoo2021} library which divides facial regions into the following parts \{\texttt{left-eye, right-eye, nose, mouth, hair, skin, background}\} (Fig.~\ref{fig:pipeline} (b)). Among the facial regions, we prioritize the following set $\mathbb{P} = \{\texttt{left-eye, right-eye, nose, mouth, hair}\}$ 
over skin and background to preserve face specific local and sparse features. 
In order to maintain pre-defined masking ratio $\mathbf{r}$, facial regions from the priority set $\mathbb{P}$ are masked across frames first followed by \texttt{\{background, skin\}} masking. Thus, Fasking generates $\mathbf{n}$ masked and ($\mathbf{k}-\mathbf{n}$) visible tokens. Across all the frames of the input $v$, we track specific facial regions from the pre-defined set to encode and reconstruct spatio-temporal changes to the model facial motion. The fasking strategy thus poses more challenges to the reconstruction while encoding subject specific appearance and fine-grained details.

\noindent \textbf{b) Masked Autoencoder.} 
After Fasking, ($\mathbf{k}-\mathbf{n}$) visible tokens are given input to the Encoder \encoder which maps the tokens to the latent space $\mathbf{z}$. The visible tokens serve as a reference to generate the masked counterpart of the face. Thus, the decoder \decoder maps the latent space $\mathbf{z}$ to the reconstructed masked tokens $X^{'}_m$. Please note that similar to VideoMAE~\cite{tongVideoMAE2022}, we adopt ViT~\cite{dosovitskiyImage2021} architecture as a backbone for MARLIN. A reconstruction loss ($\mathcal{L}_{\texttt{recon}}$) is imposed between masked cubes $X_m$ and their reconstructed counterparts $X^{'}_m$ to guide the learning objective. 

\noindent \textbf{c) Adversarial Adaptation Strategy.} To enhance the generation quality for rich representation learning, we incorporate adversarial adaptation on top of the masked auto-encoder backbone. According to the prior literature~\cite{radfordUnsupervised2016, donahueLarge2019}, adversarial training enhances generation quality which in turn results in rich latent feature $z$. The discriminator \discriminator as shown in Fig.~\ref{fig:pipeline} is an MLP based network which imposes adversarial loss $\mathcal{L}_{\texttt{adv}}$ between $X_m$ and their reconstructed counterparts $X^{'}_m$.

\subsection{Overall MARLIN Loss}
Alg.~\ref{alg:MARLIN} summarizes the training process for the MARLIN framework. MARLIN mainly imposes (a) Reconstruction Loss and (b) Adversarial Loss to facilitate the training. 

\noindent \textbf{(a) Reconstruction Loss.} Given an input masked tokens $\tilde{X}_m$, the masked auto-encoder module reconstruct it back to $X^{'}_m$. To this end, we minimize mean squared error loss in the 3D token space to update the weights of the (\discriminator $\circ$\encoder $\circ$ \fasking ) branch. The loss is defined as 

\begin{equation} 
\vspace{-2mm}
\mathcal{L}_{\texttt{recon}} = \frac{1}{N}\sum_{i=1}^{N}||X_m^{(i)} - X^{'(i)}_m||_2
\vspace{-1mm}
\end{equation} 
where $N$ is the total number of data in $\mathbb{D}$, $X_m^{(i)}$ and $X_m^{'(i)}$ are the masked token and reconstruction of $i$-th data in $\mathbb{D}$.

\noindent \textbf{(b) Adversarial Loss.} The adversarial adaptation considers the Wassenstain GAN loss~\cite{arjovskyWasserstein2017} for better reconstruction of spatio-temporal facial patterns which in turn helps in learning rich representation. The loss is defined as follows: 

\begin{align}
    \vspace{-1mm}
    \mathcal{L}^{(d)}_{\texttt{adv}} &= \frac{1}{N\mathbf{n}}\sum_{i=1}^{N}(\sum_{x^{'}_m \in X^{'(i)}_m}\mathcal{F}_{\phi_{\Gamma}}(x^{'}_m) - \sum_{x_m \in X^{(i)}_m}\mathcal{F}_{\phi_{\Gamma}}(x_m)) \label{eq:d_loss}
    \vspace{-4mm}\\
    \mathcal{L}^{(g)}_{\texttt{adv}} &= -\frac{1}{N\mathbf{n}}\sum_{i=1}^{N}\sum_{x^{'}_m \in X^{'(i)}_m}\mathcal{F}_{\phi_{\Gamma}}(x^{'}_m) \label{eq:g_loss}
    \vspace{-4mm}
\end{align}

\noindent Thus, the overall learning objective $\mathcal{L}$ is formulated as follows, where $\lambda_W$ is the weighting parameter: 

\begin{align}
    \mathcal{L}^{(g)} &= \mathcal{L}_{\texttt{recon}} + \lambda_W \mathcal{L}^{(g)}_{\texttt{adv}} \\
    \mathcal{L}^{(d)} &= \mathcal{L}^{(d)}_{\texttt{adv}}
\end{align} During MARLIN's pre-training phase, $\mathcal{L}^{(d)}$ updates the parameters $\phi_{dis}$ and $\mathcal{L}^{(g)}$ updates the parameters $\phi_{e}$, $\phi_{d}$.

\begin{algorithm}[tb]
\caption{Training procedure for MARLIN}
\label{alg:MARLIN}
\begin{algorithmic}[1]
\Require{\fasking, \encoder, \decoder, \discriminator, $\mathcal{F}_{\theta}$, $\mathbb{D}$, $\mathbf{r}$, $\mathbb{D}_{down}$}
\While{not converged} \Comment{MARLIN pre-training}
\State $v \gets \texttt{sample batch}(\mathbb{D})$
\State $\{\tilde{X}_m, \tilde{X}_v\} \gets \mathcal{F}_{\phi_{f}}(v, r)$ \Comment{Fasking (See Algo~\ref{alg:face})}
\State $X^{'}_m \gets \mathcal{F}_{\phi_{\mathcal{D}}}\circ\mathcal{F}_{\phi_{\mathcal{E}}}(\tilde{X}_v) $ \Comment{Train \discriminator}
\State $\{\phi_{\Gamma}\} \gets \triangledown_{\{\phi_{\Gamma}\}} \mathcal{L}^{(d)}(X_m, X^{'}_m) $
\State $X^{'}_m \gets \mathcal{F}_{\phi_{\mathcal{D}}}\circ\mathcal{F}_{\phi_{\mathcal{E}}}(X_v) $ \Comment{Train \encoder, \decoder}
\State $\{\phi_{\mathcal{E}}, \phi_{\mathcal{D}}\} \gets \triangledown_{\{\phi_{\mathcal{E}}, \phi_{\mathcal{D}}\}} \mathcal{L}^{(g)}(X_m, X^{'}_m)$
\EndWhile
\While{not converged} \Comment{Downstream Adaptation}
\State $\{v, \mathbf{y}\} \gets \texttt{sample batch}(\mathbb{D}_{\texttt{down}})$
\State $\tilde{X} \gets$ tokenize $v$
\State $\mathbf{y}^{'} \gets \mathcal{F}_{\theta} \circ \mathcal{F}_{\phi_{\mathcal{E}}}(\tilde{X})$ \Comment{Adapt downstream label}
\If{Linear Probing} \Comment{Linear Probing}
    \State $\{\theta\} \gets \triangledown_{\{\theta\}} \mathcal{L}_{\texttt{down}}(y, y') $
\Else \Comment{Fine-Tuning}
    \State $\{\phi_{\mathcal{E}}, \theta\} \gets \triangledown_{\{\phi_{\mathcal{E}}, \theta\}} \mathcal{L}_{\texttt{down}}(y, y') $
\EndIf
\EndWhile
\end{algorithmic}
\end{algorithm}

\subsection{Downstream Adaptation}\label{sec:downstream}

Our proposed MARLIN framework learns robust and transferable facial representation from the facial video in a self-supervised way. Following the standard evaluation protocols, we adopt Linear Probing (LP) and Fine-Tuning (FT) for downstream adaptation for different face relevant tasks (See Fig.~\ref{fig:pipeline} inference module). Given any task specific downstream dataset $\mathbb{D}_{\texttt{down}} = \{v_j,\mathbf{y}_j\}_{j=1}^{\mathcal{N}}$, we deploy linear fully-connected (FC) layers with embedding parameters $\theta$ to align the latent space to the downstream task specific label space on top of encoder module \encoder. For linear probing, we freeze the backbone network \encoder and only update the $\mathcal{F}_\theta$. On the other hand for FT, we fine-tune the whole module i.e. (\encoder $\circ \mathcal{F}_\theta$). When MARLIN is used as a feature extractor for LP, it uses a sliding temporal window to extract features $Z$ of the input face cropped video $V$ as shown in Fig.~\ref{fig:pipeline} (c). The details of different downstream facial tasks are described below:

\noindent \textbf{Facial Attribute Recognition (FAR)} predicts the presence of appearance and action attributes such as gender, race, hair color, and emotion of a given face video. The problem of predicting facial attributes can be posed as a multi-label learning problem highly dependent on rich spatial encoding. For the downstream adaptation purpose, we use 28,532 train, 3,567 val, and 3,567 test videos from the CelebV-HQ~\cite{zhuCelebVHQ2022} dataset. Following the prior works~\cite{liuDeep2015,ghoshAutomatic2018a,zhengGeneral2022}, we report average accuracy($\uparrow$), Area Under the Curve (AUC$\uparrow$) over all attributes.  

\noindent \textbf{Facial Expression Recognition (FER)} task encodes spatio-temporal facial muscle movement patterns to predict emotion (6-class) and sentiment (7-class and 2-class) of the concerned subject given a facial video. We evaluate the performance of MARLIN on CMU-MOSEI dataset~\cite{bagherzadehMultimodal2018} which is a conversational corpus having 16,726 train, 1,871 val, and 4,662 test data. Following the prior works~\cite{delbrouckTransformerbased2020,bagherzadehMultimodal2018}, we use overall accuracy($\uparrow$) as metrics.

\noindent \textbf{Deepfake Detection (DFD)} task predicts spatio-temporal facial forgery given a facial video from FF++(LQ) dataset~\cite{rosslerFaceForensics2019}. For downstream adaptation, we use 3,600 train, 700 val, and 700 test sample videos from FF++(LQ) dataset~\cite{rosslerFaceForensics2019}. Following prior literature~\cite{qianThinking2020,wangM2TR2022,caiYou2022}, we use accuracy($\uparrow$) and AUC($\uparrow$) as the evaluation metrics.

\noindent \textbf{Lip Synchronization (LS)} is another line of research that require facial region specific spatio-temporal synchronization. This downstream adaptation further elaborates the adaptation capability of MARLIN for face generation tasks. For adaptation, we replace the facial encoder module in Wav2Lip~\cite{prajwalLip2020} with MARLIN, and adjust the temporal window accordingly i.e. from 5 frames to $\mathbf{T}$ frames. For evaluation, we use the LRS2~\cite{chungLip2017} dataset having 45,838 train, 1,082 val, and 1,243 test videos. Following the prior literature~\cite{prajwalLip2020,wangAttentionBased2022}, we use Lip-Sync Error-Distance (LSE-D $\downarrow$), Lip-Sync Error-Confidence (LSE-C $\uparrow$) and Frechet Inception Distance (FID $\downarrow$)~\cite{heuselGANs2017} as evaluation matrices.

\begin{table}[t]
\caption{\textbf{Facial Attribute Recognition.} Our proposed framework, MARLIN, trained on YTF~\cite{wolfFace2011} dataset and Linear Probed/Fine-Tuned on CelebV-HQ~\cite{zhuCelebVHQ2022} benchmark dataset in terms of accuracy$\uparrow$ and area under the curve$\uparrow$. * shows supervised methods trained on the CelebV-HQ~\cite{zhuCelebVHQ2022} dataset.}
\label{tab:celebhq-v}
\scalebox{0.80}{
\begin{tabular}{l||cc|cc|c}
\toprule[0.4 mm]
\rowcolor{mygray}\textbf{ Method }  & \multicolumn{2}{c|}{\textbf{Appearance}} & \multicolumn{2}{c|}{\textbf{Action}} & \textbf{Overall}\\ \cline{2-6}
\rowcolor{mygray} & \multicolumn{1}{c|}{\textbf{Acc.}$\uparrow$}  & \multicolumn{1}{c|}{\textbf{AUC}$\uparrow$} &  \multicolumn{1}{c|}{\textbf{Acc.}$\uparrow$}    & \multicolumn{1}{c|}{\textbf{AUC}$\uparrow$}&
\multicolumn{1}{c}{\textbf{Acc.}$\uparrow$}\\ \hline \hline
R3D~\cite{tranCloser2018}* & 92.34 & 0.9424 & 94.57 & 0.9173 & 93.45\\
MViTv1~\cite{fanMultiscale2021}* & 92.90 & 0.9452 & 95.13 & 0.9233 & 94.01 \\
MViTv2~\cite{liMViTv22022}* & 92.77 & 0.954 & 95.15 & 0.9239 & 93.96 \\
VideoMAE (FT)~\cite{tongVideoMAE2022} & 92.91 & 0.9529 & 95.37 & 0.9284 & 94.14 \\\hline
MARLIN (LP) & 91.90 & 0.9373 & 95.25 & 0.9278 & 93.57 \\
MARLIN (FT) & 93.90 & 0.9561 & 95.48 & 0.9406 & 94.69 \\
\bottomrule[0.4mm]
\end{tabular}}
\vspace{-5mm}
\end{table}

\section{Experiments and Results}\label{sec:exp_results}
We have comprehensively compared our method on different downstream adaptation tasks from quantitative (See Sec.~\ref{sec:quantitative}) and qualitative (See Sec.~\ref{sec:qualitative} perspectives. Additionally, we have performed extensive ablation studies to provide justification for our design choices.
\subsection{Experimental Protocols}
\noindent \textbf{Datasets.} We evaluate the MARLIN framework on different facial analysis tasks described in Sec.~\ref{sec:downstream}. In brief, we use CelebV-HQ~\cite{zhuCelebVHQ2022} for facial attribute and action prediction, CMU-MOSEI dataset~\cite{bagherzadehMultimodal2018} for conversational emotion and sentiment prediction, FF++(LQ) dataset~\cite{rosslerFaceForensics2019} for deepfake detection and LRS2~\cite{chungLip2017} for lip synchronization.

\noindent \textbf{Settings.} For fair comparisons, we follow the dataset specific experimental protocols mentioned in the task specific prior literature~\cite{liuDeep2015,ghoshAutomatic2018a,zhengGeneral2022,rosslerFaceForensics2019,chungLip2017,bagherzadehMultimodal2018}. Other than traditional evaluation, we perform few shot adaptation strategy as well to show the robustness and transferability of MARLIN.

\noindent \textbf{Implementation Details.} We implemented the method on PyTorch~\cite{paszkePyTorch2019} with Nvidia RTX A6000 GPU. First of all, given any temporal chunk of a facial video, consecutive frames are highly redundant. Therefore, to consider semantically meaningful frames having significant motion across frames, we adopt the minimum temporal stride value to be $2$. Given an input video (having dimension $3\times 16\times 224 \times 224$), the cube embedding layer generates $8\times 14\times 14$ 3D tokens of dimension $2\times 16 \times 16$ to preserve spatio-temporal patterns. Using the Fasking strategy (See Algo.~\ref{alg:face}), MARLIN densely masks these tokens with a pre-defined masking ratio. Our empirical analysis suggests that MARLIN works favorably with a high masking ratio ($90\%$). MARLIN's objective is to generate the masked part from the sparse visible tokens. After Fasking, each token is mapped to the latent space embedding dimension of $768$. From this latent embedding, the masked part is reconstructed in the 3D token space that can further be mapped to the original video.
For fair comparison, we use ViT-B as the backbone encoder, although the impact of other ViT-variants are depicted in ablation study.
The pre-training hyperparameters are as follows: the base learning rate is linearly scaled with respect to the overall batch size, $\texttt{lr} = \texttt{base\ learning\ rate} \times \texttt{batch\ size} / \texttt{256}$. For self-supervised pre-training, we use AdamW optimizer with base learning rate $1.5e{-4}$, momentum $\beta_1 = 0.9, \beta_2 = 0.95$ with a learning rate scheduler (cosine decay)~\cite{loshchilovSGDR2022}. For linear probing, we use Adam optimizer with $\beta_1 = 0.5, \beta_2 = 0.9$ and base learning rate $1e{-4}$, weight decay 0. For fine-tuning, we use Adam optimizer with $\beta_1 = 0.5, \beta_2 = 0.9$ and base learning rate $1e{-4}$ without any weight decay. 

\begin{table}[t]
    \centering
    \caption{\textbf{Facial Expression and Sentiment Recognition.} Downstream adaptation results on MOSEI dataset~\cite{bagherzadehMultimodal2018} for Emotion, sentiment (7-class), and sentiment (2-class). Our proposed method, MARLIN, outperforms visual modality based emotion prediction methods. \textit{Please note that SOTA for UMON~\cite{delbrouckTransformerbased2020} and GMF~\cite{arevaloGated2020} utilize three modalities and thus, not directly comparable.} Here, YTF: YouTubeFace~\cite{wolfFace2011} and LAV represents linguistic, audio, and visual modality, respectively. * denotes supervised methods.} 
    \label{tab:mosei}
    \scalebox{0.85}{
    \begin{tabular}{l|c|l|c|c}
    \toprule[0.4mm]
    \rowcolor{mygray}  \textbf{Tasks} & \textbf{Pre-train} &  \textbf{Method} & \multicolumn{1}{c|}{\textbf{Mod.}}& \multicolumn{1}{c}{\textbf{Acc.}$\uparrow$}\\ \hline \hline
       \multirow{5}{*}{Emotion}
         & -- & MViTv1~\cite{liMViTv22022}* & V & 80.45 \\
         & -- & UMONS~\cite{delbrouckTransformerbased2020}* &  LAV & 80.68 \\
         & -- & GMF~\cite{arevaloGated2020}* & LAV & 81.14 \\
         & YTF~\cite{wolfFace2011} & VideoMAE~\cite{tongVideoMAE2022} & V  & 80.39 \\  
         & YTF~\cite{wolfFace2011} & MARLIN & V & 80.60 \\\hline
        \multirow{3}{*}{\begin{tabular}[c]{@{}c@{}}Sentiment\\ (7-Class)\end{tabular}}  
         & -- & MViTv1~\cite{liMViTv22022}* & V & 33.35 \\
        & YTF~\cite{wolfFace2011} & VideoMAE~\cite{tongVideoMAE2022} & V & 33.78 \\
        & YTF~\cite{wolfFace2011} & MARLIN & V & 34.63 \\\hline
         \multirow{4}{*}{\begin{tabular}[c]{@{}c@{}}Sentiment\\ (2-Class)\end{tabular}}  
         & MOSEI~\cite{bagherzadehMultimodal2018} and & \multirow{2}{*}{CAE-LR~\cite{koromilasUnsupervised2022}} & \multirow{2}{*}{V} & \multirow{2}{*}{71.06} \\
         & IEMOCAP~\cite{bussoIEMOCAP2008} & & & \\
        & YTF~\cite{wolfFace2011} & VideoMAE~\cite{tongVideoMAE2022} & V & 72.96 \\
        & YTF~\cite{wolfFace2011} & MARLIN & V &  73.70 \\
        \bottomrule[0.4mm]
\end{tabular}}
\vspace{-5mm}
\end{table}

\begin{table}[b]
\vspace{-4mm}
\centering
\caption{\textbf{Deepfake Detection.} We compare the 
Fine-Tuning (FT) results on MARLIN for FaceForensic++~\cite{rosslerFaceForensics2019} dataset. * denotes supervised methods.}
\label{tab:ff++}
\scalebox{0.85}{
    \begin{tabular}{c|l|cc}
    \toprule[0.4mm]
    \rowcolor{mygray} \textbf{Pre-train} &  \textbf{Method} &  \multicolumn{1}{c}{\textbf{Acc.(\%)}$\uparrow$} & \multicolumn{1}{c}{\textbf{AUC}$\uparrow$}\\  \hline \hline
    -- & Steg.Features~\cite{fridrichRich2012}* & 55.98 & --\\
    -- & LD-CNN~\cite{cozzolinoRecasting2017}* & 58.69 & --\\
    -- & Constraied Conv.~\cite{bayarDeep2016}* & 66.84  & --\\
    -- & CustomPooling CNN~\cite{rahmouniDistinguishing2017}*  & 61.18 & --\\
    -- & MesoNet~\cite{afcharMesoNet2018}* & 70.47 & -- \\
    -- & Face X-ray~\cite{liFace2020}* & -- & 0.6160\\
    -- & Xception~\cite{cholletXception2017}*  & 86.86 & 0.8930 \\
    -- & F$^3$-Net~\cite{qianThinking2020}* & 93.02 & 0.9580 \\
    -- & P3D~\cite{qiuLearning2017}*  &-- & 0.6705 \\
    -- &R3D~\cite{tranCloser2018}*  &  --& 0.8772 \\
    -- &I3D~\cite{carreiraQuo2017}*&--& 0.9318 \\
    -- &M2TR~\cite{wangM2TR2022}* &  -- & 0.9395 \\
    -- &ST-M2TR~\cite{wangM2TR2022}*& --& 0.9531 \\     
    YTF~\cite{wolfFace2011} & VideoMAE~\cite{tongVideoMAE2022} & 87.57 &  0.9082\\\hline
    YTF~\cite{wolfFace2011} & MARLIN &  89.43  & 0.9305\\
    \bottomrule[0.4mm]
    \end{tabular}}
\end{table}

\begin{table}[t]
\caption{\textbf{Lip Synchronization.} We compare Linear Probing (LP) and Fine-Tuning (FT) results on the LRS2~\cite{chungLip2017} dataset.}
\label{tab:LRS2}
\scalebox{0.80}{
\begin{tabular}{l||ccc}
\toprule[0.4mm]
\rowcolor{mygray}  \textbf{Method} & \textbf{LSE-D}$\downarrow$ & \textbf{LSE-C }$\uparrow$ & \textbf{FID}$\downarrow$ \\\hline \hline
Speech2Vid~\cite{jamaludinYou2019}               & 14.230                           & 1.587                           & 12.320                         \\
LipGAN~\cite{krAutomatic2019}                  & 10.330                           & 3.199                           & 4.861                         \\
Wav2Lip~\cite{prajwalLip2020}                  & 7.521                           & 6.406                           & 4.887                         \\
AttnWav2Lip~\cite{wangAttentionBased2022}        & 7.339                           & 6.530                            & --          \\
Wav2Lip + ViT~\cite{dosovitskiyImage2021}            & 8.996                           & 2.807                           & 13.352                        \\
Wav2Lip + ViT + VideoMAE~\cite{tongVideoMAE2022} & 7.316                           & 5.096                           & 4.097                         \\
Wav2Lip + ViT + MARLIN   & 7.127                           & 5.528                           & 3.452  \\
\bottomrule[0.4mm]
\end{tabular}}
\vspace{-6mm}
\end{table}

\subsection{Quantitative Analysis}\label{sec:quantitative}

\noindent \textbf{4.2.1. Comparison with SOTA Facial Analysis Tasks.}\\
We compare the performance of MARLIN with different downstream facial analysis tasks following standard task specific evaluation protocols~\cite{liuDeep2015,ghoshAutomatic2018a,zhengGeneral2022,rosslerFaceForensics2019,chungLip2017,bagherzadehMultimodal2018}. 

\noindent \textbf{Facial Attributes.} In Tab.~\ref{tab:celebhq-v}, we compare the LP and FT adaptation performance of MARLIN with the popular transformer (i.e. MViT-v1~\cite{fanMultiscale2021} and MViT-v2~\cite{liMViTv22022}) and CNNs (i.e. R3D~\cite{tranCloser2018}) on CelebV-HQ~\cite{zhuCelebVHQ2022} dataset. From the table, it is observed that MARLIN's FT version outperforms supervised MViT-v2~\cite{liMViTv22022} transformer architecture by 1.13\% (92.77\% $\rightarrow$ 93.90\%) for appearance attributes and 0.33\% (95.15\% $\rightarrow$ 95.48\%) for action attributes. Similar patterns are also been observed with the R3D CNN module as well. We attribute MARLIN's performance gain to the pre-training strategy that encodes generic, robust, and transferable features from any input facial video.

\noindent \textbf{Emotion and Sentiment.} In Tab.~\ref{tab:mosei}, we similarly compare the LP and FT adaptation performance of conversational emotion and sentiment in terms of accuracy($\uparrow$) and AUC($\uparrow$) on CMU-MOSEI~\cite{zadehMultiattention2018a} dataset. \textit{Please note that the MARLIN is a visual modality only encoder.} The results suggest that MARLIN performs competitively with SOTA methods~\cite{koromilasUnsupervised2022,liMViTv22022,delbrouckTransformerbased2020}, especially it outperforms unsupervised SOTA CAE-LR~\cite{koromilasUnsupervised2022} by 2.64\% (71.06\% $\rightarrow$ 73.70\%) on 2-class sentiment task. For emotion and 7-class sentiment as well, it outperforms supervised benchmarks~\cite{liMViTv22022} marginally. These results also indicate that MARLIN learns highly generic, robust, and transferable feature representation from pre-training.

\noindent \textbf{DeepFake Detection.} In Tab.~\ref{tab:ff++}, we compare the performance of video manipulation on FaceForensics++~\cite{rosslerFaceForensics2019} dataset and report results in terms of video-level accuracy($\uparrow$) and AUC($\uparrow$). The results indicate that MARLIN performs favorably against the supervised SOTA methods~\cite{afcharMesoNet2018,fridrichRich2012,cozzolinoRecasting2017,bayarDeep2016,rahmouniDistinguishing2017,liFace2020,cholletXception2017,qiuLearning2017,tranCloser2018,carreiraQuo2017}. This is the first SSL work that uses only spatio-temporal visual information anomaly to detect video manipulation. Unless F$^3$-Net, which uses frequency aware pattern over the temporal dimension to detect forgeries in a supervised fashion. Whereas MARLIN irrespective of frequency pattern learns facial representation and can detect anomalies from the spatio-temporal signal.

\noindent \textbf{Lip Synchronization.} For a fair comparison, we adopt the following experimental setups: \textit{1) Wav2Lip+ViT:} To compare the contribution of ViT architecture~\cite{dosovitskiyImage2021} wrt SOTA CNNs and MARLIN where the weights of ViT is trained from scratch on LRS2~\cite{chungLip2017} dataset. \textit{2) Wav2Lip+ViT+VideoMAE:} To compare the contribution of vanilla VideoMAE with ViT backbone pre-trained on YTF~\cite{wolfFace2011} dataset. \textit{2) Wav2Lip+ViT+MARLIN:} To compare the contribution of MARLIN pre-trained on YTF~\cite{wolfFace2011} with SOTA~\cite{siSpeech2Video2021,prajwalLip2020,wangAttentionBased2022} and different design aspects. The experimental results are depicted in Tab.~\ref{tab:LRS2} with LSE-D$\downarrow$, LSE-C $\uparrow$ and FID $\downarrow$ as evaluation metrics following standard protocol~\cite{heuselGANs2017,siSpeech2Video2021,prajwalLip2020,wangAttentionBased2022}. The improvement of lip sync score (LSE-D$\downarrow$: 7.521 $\rightarrow$ 7.127; FID $\downarrow$: 4.887 $\rightarrow$ 3.452) indicates that MARLIN learns rich spatio-temporal patterns which are transferable and robust. It is also interesting to observe that MARLIN is adaptive to very fine grained features specific to the face as well. 

\begin{table}[t]
\centering
\caption{\textbf{Few shot adaptation on different facial tasks.} Comparison of different methods for few shot adaptation.}
\scalebox{0.80}{
\begin{tabular}{c||ccc|c|cc}
\toprule[0.4mm]
\rowcolor{mygray}\textbf{Data}$\rightarrow$ & \multicolumn{3}{c|}{\textbf{MOSEI~~\cite{bagherzadehMultimodal2018}}} & \multicolumn{1}{c|}{\textbf{FF++~\cite{rosslerFaceForensics2019}}} & \multicolumn{2}{c}{\textbf{CelebV-HQ~\cite{zhuCelebVHQ2022}}} \\ \cline{2-7}
\rowcolor{mygray}\textbf{Task}$\rightarrow$ & \textbf{Emo.} & \textbf{7-Sen.} & \textbf{2-Sen.} & \multicolumn{1}{c|}{\textbf{DeepFake}} & \multicolumn{1}{c}{\textbf{Appr.}} & \multicolumn{1}{c}{\textbf{Act.}} \\ 
\rowcolor{mygray}\textbf{Anno.\%} & \textbf{Acc.}$\uparrow$ & \textbf{Acc.}$\uparrow$ & \textbf{Acc.}$\uparrow$ & \textbf{AUC}$\uparrow$ & \textbf{AUC}$\uparrow$ &  \textbf{AUC}$\uparrow$ \\\hline \hline
100\% & 80.60 & 34.63 & 73.70 & 0.9305 & 0.9373 & 0.9278  \\ 
50\% & 80.59 & 33.73 & 73.33 & 0.8681  & 0.9273 & 0.9270 \\
10\% & 79.89 & 33.56 & 72.26 & 0.7459 & 0.8996 & 0.9201  \\ 
1\% & 78.61 & 30.09 & 71.89 & 0.6252 & 0.8423 & 0.9063  \\ 
\bottomrule[0.4mm]
\end{tabular}}
\label{tab:few-shot}
\vspace{-5mm}
\end{table}

\noindent \textbf{4.2.2. Few-Shot Adaptation.}\\
Few shot adaptation has recently gained attention due to its adaptation capability with very low data regime~\cite{browatzki3FabRec2020,shuLearning2021,zhuangFewShot2021,zhengGeneral2022}. Following the standard evaluation protocol~\cite{browatzki3FabRec2020,shuLearning2021,zhuangFewShot2021,zhengGeneral2022}, we also investigate the adaptation capability of MARLIN. Given any downstream dataset, we use limited train set labels to align the output manifold while keeping the test set fixed via LP (MOSEI, CelebV-HQ) and FT (FF+) strategy. From Tab.~\ref{tab:few-shot}, a slight drop in performance is observed across different tasks which further demonstrates that MARLIN learns generic, transferable, and adaptive information. 

\begin{table}[b]
\vspace{-4mm}
\caption{Contribution of different modules, encoder architectures, and masking strategies towards overall MARLIN framework. Fasking: Facial Guided Masking, AT: Adversarial Training}
\label{tab:ablation}
\scalebox{0.80}{
\begin{tabular}{l||ccc||cc}
\toprule[0.4mm]
\rowcolor{mygray} \textbf{Datasets} $\rightarrow$ & \multicolumn{3}{c||}{\textbf{MOSEI~\cite{bagherzadehMultimodal2018}}} & \multicolumn{2}{c}{\textbf{FF++~\cite{rosslerFaceForensics2019}}}\\
\rowcolor{mygray} & \begin{tabular}[c]{@{}c@{}}\textbf{Emo.}\\ Acc. \\(\%$\uparrow$) \end{tabular}& \begin{tabular}[c]{@{}c@{}}\textbf{7-Sent.}\\ Acc. \\ (\%$\uparrow$) \end{tabular}& \begin{tabular}[c]{@{}c@{}}\textbf{2-Sent.}\\ Acc. \\ (\%$\uparrow$) \end{tabular}& \begin{tabular}[c]{@{}c@{}}\\\textbf{Acc.}\\ (\%$\uparrow$) \end{tabular}& \begin{tabular}[c]{@{}c@{}}\\\textbf{AUC.}\\ ($\uparrow$) \end{tabular} \\ \hline \hline
\textbf{Modules} $\downarrow$  & & & & &  \\
VideoMAE        & 80.39                                                                     & 33.78               & 72.96               & 87.57                                                    & 0.9082               \\
+ Fasking & 80.55                                                                     & 34.58               & 73.54               &  87.29                                                    & 0.9154               \\
+ AT  & 80.58                                                                     & 34.05               & 73.17               & 88.00                                                      & 0.9096               \\
+ Both (MARLIN)         & \textbf{80.60}                                                                      & \textbf{34.63}               & \textbf{73.70}                 & \textbf{89.43}                                                    & \textbf{0.9305}   \\  \hline \hline
\textbf{Encoder Arch.} $\downarrow$ & & & & &  \\ 
ViT-S & 80.38 & 33.40 & 72.69 & 87.43 & 0.8863 \\
ViT-B & 80.60 & 34.63 & 73.70 & 89.43 & 0.9305 \\  
ViT-L & \textbf{80.63} & \textbf{35.28} & \textbf{74.83} & \textbf{90.71} & \textbf{0.9377} \\ \hline \hline
\textbf{Masking Strategy} $\downarrow$ & & & & &  \\ 
Random & 80.40 & 34.10 & 72.96 & 87.29 & 0.8797 \\
Frame & 79.33 & 33.99 & 72.90 & 86.57 & 0.8835 \\
Tube & 80.58 & 34.05 & 73.17 & 88.00 & 0.9096 \\
Fasking & \textbf{80.60} & \textbf{34.63} & \textbf{73.70} & \textbf{89.43} & \textbf{0.9305} \\   
\bottomrule[0.4mm]
\end{tabular}}
\end{table}

\begin{figure}[t]
    \centering
    \includegraphics[width = \linewidth]{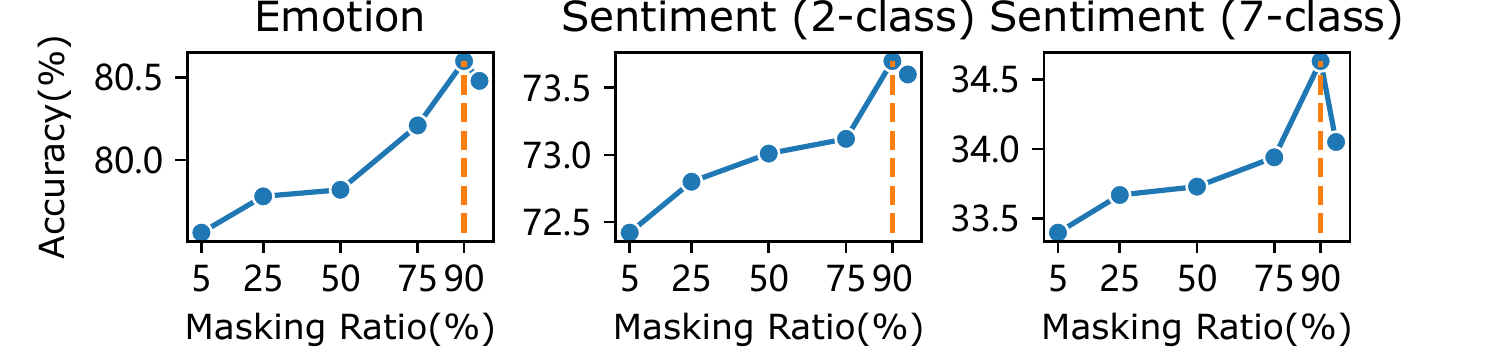}
    \caption{\textbf{Impact of Masking Ratio} Comparison of different masking ratios for emotion and sentiment prediction in CMU-MOSEI dataset~\cite{bagherzadehMultimodal2018}. Empirically, it suggests $90\%$ masking works best for MARLIN.
    }
    \label{fig:masking_ratio}
    \vspace{-6mm}
\end{figure}

\noindent \textbf{4.2.3. Ablation Studies.}\\
We have performed extensive ablation studies to show the effectiveness of each component.

\noindent \textbf{1) Masking ratio.}
We use different masking ratios in the range [0.05 - 0.95] and repeat the pre-training followed by LP on CMU-MOSEI~\cite{zadehMultiattention2018a} dataset. From Fig.~\ref{fig:masking_ratio}, we see that $\sim 90\%$ masking ratio is optimal for the MARLIN. With a less masking ratio (i.e. $\leq$ 0.5 ), more information is available for the reconstruction task which degrades the feature quality. Similarly, beyond $\sim 90\%$, the reconstruction task becomes more challenging, leading to a performance drop. With the empirical evidence, we set the masking ratio to be $\sim 90\%$ throughout all of our experiments.
\textbf{2) Masking strategies.}
We further compare the proposed \textit{Fasking} strategy with existing masking strategies~\cite{tongVideoMAE2022,feichtenhoferMasked2022} \ie \textit{Frame}, \textit{Random} and \textit{Tube-Masking}. The empirical results in Tab.~\ref{tab:ablation} demonstrate that \textit{Fasking} is better.
\textbf{3) Different modules.} We progressively integrate each module and observe its impact on downstream performance on CMU-MOSEI~\cite{zadehMultiattention2018a} and FF++~\cite{rosslerFaceForensics2019} while keeping other components fixed. From Tab.~\ref{tab:ablation}, we see that the addition of Fasking and Adversarial Training (AT) improves the performance, reflecting the importance of each component. 
\textbf{4) Encoder architectures.} To investigate the impact of the backbone encoder architectures, and compare ViT-S, ViT-B, and ViT-L (See Tab.~\ref{tab:ablation}). We observe that the larger model size enhances the performance. For fair comparison, we use a ViT-B encoder.

\vspace{-1mm}
\subsection{Qualitative Aspects}\label{sec:qualitative}
\vspace{-1mm}
\noindent In order to understand the effectiveness of the learned features, we further conducted following qualitative analysis.

\begin{figure}
    \centering
    \includegraphics[width=\linewidth]{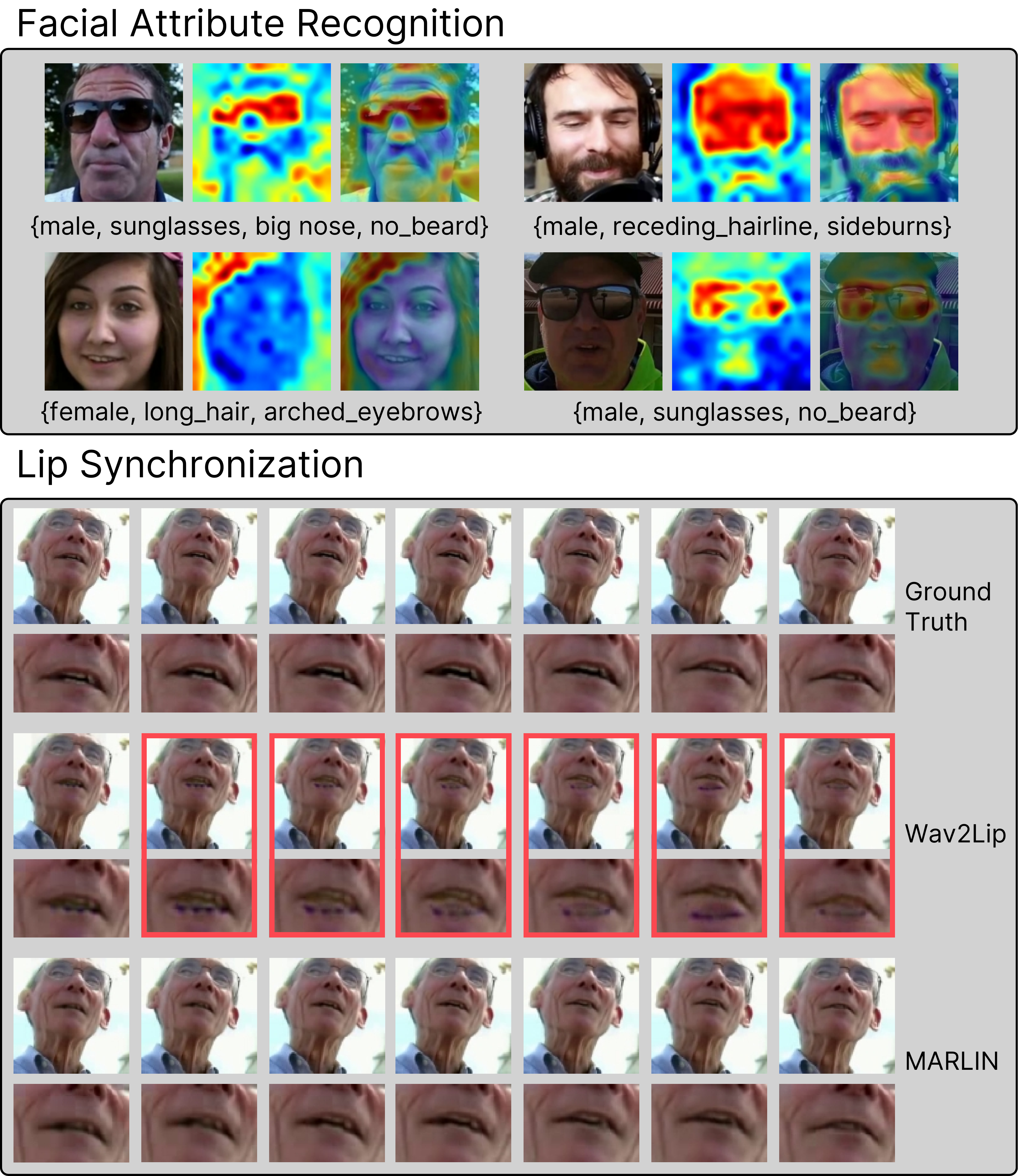}
    \caption{\textbf{Qualitative Analysis.} \textit{Top:} Qualitative results for MARLIN for facial attribute recognition task. \textit{Bottom:} Qualitative results for MARLIN for facial lip synchronization task.}
    \label{fig:grad_cam}
    \vspace{-7mm}
\end{figure}

\noindent \textbf{1) Facial Attributes.} We visualize the important regions that MARLIN focused on using Gradient-weighted Class Activation Mapping (Grad-CAM)~\cite{selvarajuGradCAM2017}. In Fig.~\ref{fig:grad_cam} top, the heat-map results are based on LP on top of MARLIN's feature on CelebV-HQ~\cite{zhuCelebVHQ2022} dataset (appearance task) and it indicates that MARLIN focus on facial attributes such as hair, spectacle, hat, etc. 
\textbf{2) Lip Synchronization.} In Fig.~\ref{fig:grad_cam} bottom, we presents the generation results for lower part of faces which is a challenging task. The top, middle and bottom rows show ground truth, vanilla Wav2Lip~\cite{prajwalLip2020}'s output and MARLIN's output along with the closeup looks, respectively. Here, Wav2lip's CNN encoder failed to locate the lip region (as shown in the Wav2lip row of Fig.~\ref{fig:grad_cam} highlighted in red) whereas MARLIN despite pre-trained on fasking strategy is adaptive enough to generate more accurate spatio-temporal pattern for LS. 

\vspace{-1mm}
\section{Conclusion}
\vspace{-1mm}
In this paper, we aim to learn a universal and generic facial encoder, MARLIN, which is adaptive, robust and transferable for different facial analysis tasks. As a challenging auxiliary task, MARLIN reconstructs the spatio-temporal details of the face from the densely masked facial regions to capture local and global aspects which in turn helps in encoding generic and transferable features. 
\textbf{Broader Impact.} We believe that MARLIN can act as a good feature extractor for different downstream facial analysis tasks. Owing to the rich facial features, it would be easy to deploy MARLIN in low resource (e.g. mobile devices, Jetson Nano platforms) devices for real world applications. \textbf{Limitations.} As the model is trained on YouTube Face dataset~\cite{wolfFace2011}, there could be potential bias in terms of race and cultural background of the identities. Potential bias can also be introduced in the model as we use the existing face detection library~\cite{wangFaceXZoo2021}. We will eliminate these limitations in our updated versions.

\noindent \textbf{Acknowledgements.}\\ 
This material is based on research sponsored by DARPA under agreement number HR001122C0029. The U.S. Government is authorized to reproduce and distribute reprints for Governmental purposes notwithstanding any copyright notation thereon. M. Hayat is supported by the ARC DECRA Fellowship DE200101100.

{\small
\bibliographystyle{ieee_fullname}
\bibliography{egbib}
}

\end{document}